\pdfoutput=1
\documentclass[11pt]{article}

\usepackage{acl}
\usepackage{times}
\usepackage{latexsym}
\usepackage{subcaption}
\usepackage[T1]{fontenc}

\usepackage[utf8]{inputenc}

\usepackage{microtype}
\usepackage{afterpage}

\usepackage{inconsolata}

\usepackage{graphicx}
\usepackage{booktabs}
\usepackage{graphicx}

\usepackage{textcomp} 
\usepackage{amsmath}
\usepackage{enumitem}

\usepackage{multirow}
\usepackage{booktabs}
\usepackage{float}

\usepackage{multirow}
\usepackage{tabularx}
\usepackage{booktabs}
\usepackage{float}
\usepackage{hyperref}

\usepackage{titlesec}
\titlespacing*{\section}
{0pt}{*0.6}{*0.6}

\title{Document-level Clinical Entity and Relation Extraction via Knowledge Base-Guided Generation}

 \author{Kriti Bhattarai\textsuperscript{1,2}, \ Inez Y. Oh\textsuperscript{1}, \ Zachary B. Abrams\textsuperscript{1}, \ Albert M. Lai\textsuperscript{1,2} \\
Institute for Informatics, Data Science \& Biostatistics, Washington University School of Medicine\\  Department of Computer Science, Washington University in St. Louis}

\begin{document}
\maketitle
\begin{abstract}


Generative pre-trained transformer (GPT) models have shown promise in clinical entity and relation extraction tasks because of their precise extraction and contextual understanding capability. In this work, we further leverage the Unified Medical Language System (UMLS) knowledge base to accurately identify medical concepts and improve clinical entity and relation extraction at the document level. 
Our framework selects UMLS concepts relevant to the text and combines them with prompts to guide language models in extracting entities. Our experiments demonstrate that this initial concept mapping and the inclusion of these mapped concepts in the prompts improves extraction results compared to few-shot extraction tasks on generic language models that do not leverage UMLS. Further, our results show that this approach is more  effective than the standard Retrieval Augmented Generation (RAG) technique, where retrieved data is compared with prompt embeddings to generate results.  Overall, we find that integrating UMLS concepts with GPT models significantly improves entity and relation identification, outperforming the baseline and RAG models. By combining the precise concept mapping capability of knowledge-based approaches like UMLS with the contextual understanding capability of GPT, our method highlights the potential of these approaches in specialized domains like healthcare.

\end{abstract}
\section{Introduction}
Generative pre-trained transformer (GPT) models have shown significant potential across various clinical tasks, including information extraction, summarization, and question-answering (\citealp{Agrawal:22}; \citealp{Tang:23}; \citealp{Yang:22}, \citealp{Singhal:23}). Generative models are able to generate contextually relevant text given a prompt. However, for real-world clinical use, in tasks that require high precision, it is equally important to understand the context and minimize the errors that come from GPT models. However, accuracy of these models is limited to their training data. While GPT models are great at capturing nuanced contextual information, they often fall short in accurately identifying all medical concepts, possibly due to limited or outdated domain-specific data (\citealp{Tang:23b}, \citealp{Singhal:23}). \newline 
\indent Knowledge bases store domain-specific data. Medical knowledge bases, such as, Unified Medical Language System (UMLS) knowledge base (\citealp{Bodenreider:04}), include comprehensive information about medical concepts. Integrating knowledge bases with language  models is an open research area with multiple works exploring different ways of integrating them with language models, such as BERT (\citealp{Devlin:19}). There are limited studies on the integration of medical knowledge bases, particularly UMLS, with most recent large language models (LLMs), such as GPT.\newline 
\indent To address this limitation, we introduce an approach for clinical entity extraction that leverages UMLS for knowledge augmentation. While GPT can identify nuanced contextual information, UMLS includes a comprehensive repository of domain-specific clinical concepts that GPT may not recognize, such as brand names for drugs, abbreviations, acronyms, and aliases (\citealp{Agrawal:22}). 
\newline
\indent Our contributions in this paper are summarized as follows:\newline (1) we introduce a framework to integrate UMLS concepts into the default generative models to facilitate few-shot information extraction of biomedical entities and relations. \newline
(2) we explore current state-of-the-art knowledge augmentation techniques, such as Retrieval Augmented Generation (RAG) aimed at improving extraction, and \newline
(3) we conduct evaluation of our framework, comparing the performance of models augmented with UMLS knowledge with and without RAG, and against those without augmentation.

\section{Related Work}
\subsection{Few-shot in-context learning}

With the introduction of GPT models, there have been several works around few-shot in-context learning for clinical entity extraction where prompts guide information extraction in a contextually relevant manner (\citealp{Agrawal:22}; \citealp{Hu:24}; \citealp{Shyr:24}, \citealp{Brown:20}). Generative models can provide nuanced contextual understanding to extract clinical concepts, but cannot identify all domain-specific terminologies, especially in the clinical domain (\citealp{Tang:23b}). While recent language models have demonstrated improvement over prior language models (\citealp{Guevara:24}), there remains room for performance improvement. 

\subsection{Knowledge base-guided models}
Previous research has explored the integration of knowledge bases to enhance information extraction tasks. (\citealp{Sastre:20}) proposed a Bi-LSTM model to identify drug-related information and integrate it into knowledge graph embeddings to evaluate drug identification accuracy.  (\citealp{Gilbert:24}) addressed how knowledge bases complement language models for medical information identification tasks. Recently, a RAG model, Almanac, demonstrated significant performance improvements compared to the standard LLMs across various metrics (\citealp{Zakka:24}), further showing the benefits of access to domain-specific corpora for information extraction. 

\section{Methods}

\begin{figure}[htbp] 

  \includegraphics[width=0.5\textwidth]{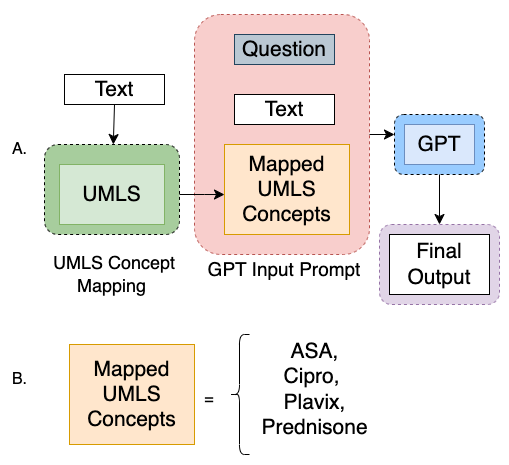}
  \caption{(A) Step-by-step approach to integrating UMLS and extracting relation pairs. (B) Example of UMLS
concepts mapped from the text. Some of the concepts, such as Prednisone, are recognized by GPT, as they are
concepts GPT model is inherently trained on. However, concepts such as ASA, Cipro, Plavix are not recognized by
GPT; UMLS facilitates their recognition.}
  \label{fig:example-pdf}

\end{figure}

\subsection{Overview of the Framework}
Our approach leverages the context-capturing capability of GPT and knowledge-capturing capability of UMLS. UMLS contains a comprehensive list of more than 1 million biomedical concepts from over 100 source vocabularies. By using the concepts in the prompts in a few-shot learning setting, we attempt to improve GPT’s ability to identify entities with the specified context that it may otherwise fail to extract independently. We map UMLS concepts to each text instance to create dynamic prompts unique to the specific context of the clinical text.The overview of the proposed framework is displayed in Figure 1.  

\subsection{UMLS Integration in Large Language
Model}\vspace{-6pt}\noindent \textit{UMLS Concept Mapping}\newline
 We first map UMLS concepts from clinical text using MetaMap (\citealp{Aronson:01}).
Given clinical text \( X = \{x_1, x_2, \ldots, x_n\} \) where \( x_i \) represents the \( i \)th clinical text, we map \( C_i = \{c_{i1}, c_{i2}, \ldots, c_{in}\} \), where \( C_i \) denote the set of concepts identified by MetaMap from \( x_i \). \( n \) denotes the number of concepts identified from \( x_i \).
These concepts are extracted leveraging MetaMap's lexical parsing, syntactic  analysis, semantic mapping, and concept mapping techniques.\newline
\indent Next, we filter the mapped concepts to include only those categorized as 'organic chemical', 'antibiotic', or 'pharmacologic substance' within the UMLS concept hierarchy as these groups contains the medications. For this work, we only target and filter medication-related concepts for augmentation and for further analysis. Let's denote the filtered set of concepts for the $i$th input clinical text $x_i$ as $C_{{filtered}}$, such that $C_{filtered,i} = \{c_{ij} \in C_i | c_{ij} \in \text{filtered groups}\}$. 
\begin{figure}[H] 
  \includegraphics[width=0.5\textwidth]{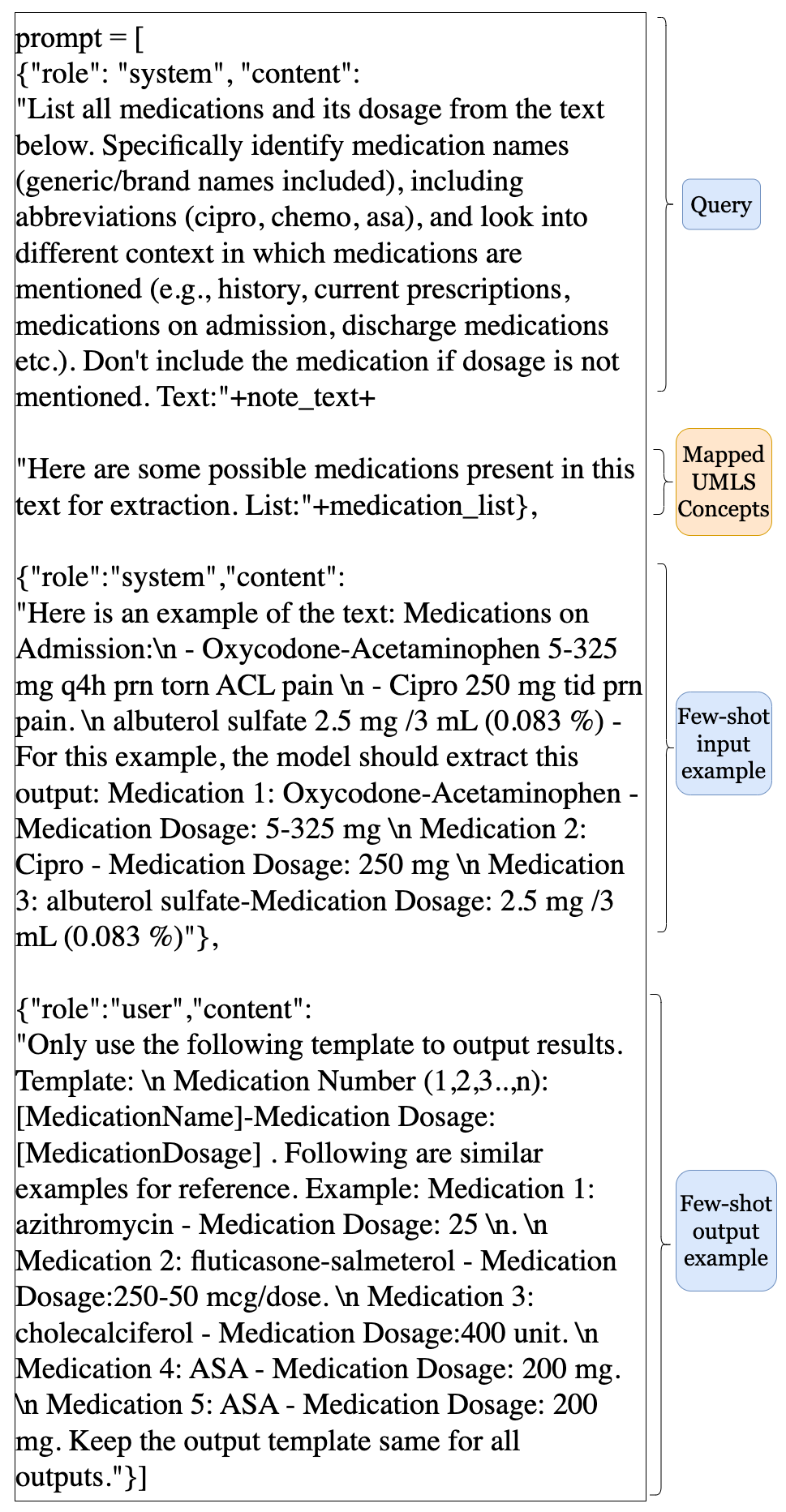}
  \caption{An example of a prompt used to extract dosage information from the text using the UMLS concepts. The \textquoteleft{note\_text}\textquoteright  \, represents each text instance from ADE or n2c2 corpus. The \textquoteleft{medication\_list}\textquoteright  \, represents the UMLS concepts extracted from MetaMap.}
  \label{fig:example-pdf}
\end{figure}

\noindent \textit{Prompt Strategy and Large Language Model Implementation}\newline
Next, we prompt the GPT model to extract entity-relation pairs from the text, leveraging the mapped UMLS concepts from MetaMap, and employing a few-shot prompt strategy. Let $P_i$ represent the prompt generated for each input text $x_i$, incorporating the relevant UMLS concepts $C_{filtered,i}$. The final prompt $P_i$ is constructed as the concatenation of the initial prompt and the set of UMLS concepts, i.e., $P_i = \text{Concat}(P, C_{filtered,i})$. We use OpenAI's GPT-4-32k (Version 0613) and GPT-3.5-turbo (Version 0301) via  HIPPA-compliant Microsoft Azure's OpenAI REST API\footnote{\url{https://learn.microsoft.com/en-us/azure/ai-services/openai/quickstart?tabs=command-line\%2Cpython-new\&pivots=programming-language-python}}endpoint.
 A sample prompt and hyperparameters used by the models for this task are available in Figure 2 and \ref{subsec:hyperparameters} respectively.  As our goal for the project was not to explore different prompting strategies, we tested a few prompts and selected the prompt that generated more specific result. We used the same format for all relation pairs replacing only the entity type for every run.\newline

\noindent \textit{Retrieval Augmented Generation}\newline
\indent We also explored another approach-RAG to leverage UMLS in a language model, which is a more conventional method involving the use of external data. RAG was chosen for its potential to enhance the generation process by incorporating domain-specific knowledge from sources like the UMLS knowledge base. Appendix \ref{subsec:RAG} includes details on our RAG implementation. \newline

\subsection{Datasets Description}
We used the n2c2 and ADE datasets for our experiments. 

\noindent \textit{n2c2 Dataset}\newline
\indent We used a curated National NLP Clinical Challenges (n2c2) dataset (\citealp{Henry:19}) consisting of 303 deidentified discharge summaries obtained from the MIMIC-III (Medical Information Mart for Intensive Care-III) critical care database (\textbf{Table 1A}) (\citealp{Johnson:16}). The data also contained annotations of medication-related entities and their relationship to other entities. Annotations conducted by 3 subject matter experts served as a gold standard to evaluate model performance.\newline

\noindent \textit{ADE Dataset}\newline
\indent The Adverse Drug Events (ADE) dataset  annotated by 5 individuals consists of MEDLINE\footnote{\url{https://www.nlm.nih.gov/medline/medline_home.html}} case reports with information on medications, dosages and adverse effects associated with the medications (\citealp{Gurulingappa:12}) (\textbf{Table 1B}). It also contains relationships between medications, dosages, and adverse effects. For our experiments, we used the second version of the dataset downloaded from Huggingface\footnote{\url{https://huggingface.co/datasets/ade_corpus_v2}}.
 
\begin{figure}[htbp] 
  
  \includegraphics[width=0.5\textwidth]{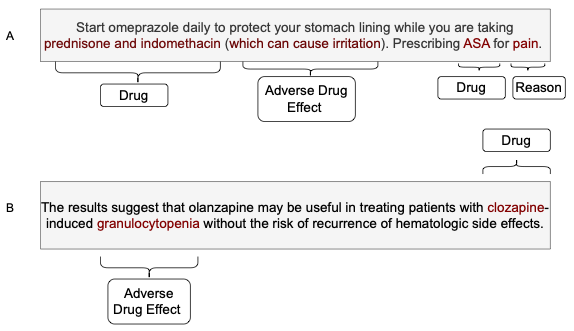}
  \caption{Sample text of discharge summaries in the (A) n2c2 dataset and (B) ADE Corpus. The text highlighted in red are the targeted entities for extraction}
  \label{fig:example-pdf}
\end{figure}
\begin{table}[htbp]
  \centering
  \caption{Statistics on the relation pairs in the (A) n2c2 dataset and the (B) ADE dataset}
  \begin{subtable}{\linewidth}
    \centering
    \begin{tabular}{llr}
      \toprule
      \multicolumn{2}{c}{\textbf{A. n2c2 Dataset}} \\
      \midrule
      \textbf{Entity-Entity Relation} & \textbf{Total instances} \\
      \midrule
      Strength-Drug     & 13338 \\
      Duration-Drug     & 643 \\
      Route-Drug        & 11038 \\
      Form-Drug         & 6636 \\
      ADE-Drug          & 2214 \\
      Dosage-Drug       & 4207 \\
      Reason-Drug       & 5160 \\
      Frequency-Drug    & 6288 \\
      \bottomrule
    \end{tabular}
  \end{subtable}
  \begin{subtable}{\linewidth}
    \centering
    \begin{tabular}{llr}
      \toprule
      \multicolumn{2}{c}{\textbf{B. ADE Dataset}} \\
      \midrule
      \textbf{Entity-Entity Relation} & \textbf{Total instances} \\
      \midrule
      Drug-ADE     & 6821 \\
      Drug-Dosage             & 279 \\
      \bottomrule
    \end{tabular} 
  \end{subtable} 
\end{table}

\begin{table*}[htbp]
\centering
\caption{Comparison of models on n2c2 Dataset and ADE Corpus. The reported results are micro-averaged precision, recall, and F-1 scores across all entity-entity relation pairs within the datasets.}
\label{tab:model-comparison}
\begin{tabular}{@{}lcccccccccccc@{}}
\toprule
\multirow{3}{*}{\textbf{Model}} & \multicolumn{3}{c}{\textbf{n2c2 Dataset}} & & \multicolumn{3}{c}{\textbf{ADE Corpus}} \\ \cmidrule(lr){2-4} \cmidrule(lr){6-8} 
 & \textbf{Precision} & \textbf{Recall} & \textbf{F1} & & \textbf{Precision} & \textbf{Recall} & \textbf{F1} \\ \midrule
GPT-3.5-turbo &  0.73 & 0.74 & 0.73 & & 0.625 & 0.57 & 0.596 \\ \midrule
GPT-3.5-turbo + UMLS & 0.77 & 0.77 & 0.77 & & 0.83 & 0.70 & 0.75 \\ \midrule
RAG w/ GPT-3.5-turbo & 0.73 & 0.74 & 0.74 & & 0.65 & 0.63 & 0.64\\ \midrule
GPT-4-32k & 0.75 & 0.76 & 0.76 & & \textbf{1.0} & 0.70 & 0.82\\ \midrule
GPT-4-32k + UMLS & \textbf{0.79} & \textbf{0.80} & \textbf{0.80} & & \textbf{1.0} & \textbf{0.89} & \textbf{0.94} \\ \midrule 
RAG w/ GPT-4-32k & 0.77 & 0.76 & 0.76 & & \textbf{1.0} & 0.74 & 0.85\\ \bottomrule
 
\end{tabular}
\end{table*}


\section{Results}

\subsection{Experimental Setup}
We evaluated two generative models, GPT-4-32k and GPT-3.5-turbo, with and without UMLS integration, and the RAG model to access the quality of generated outputs. All models were evaluated against the gold-standard annotations using precision, recall, and micro-F1 score. 

\subsection{Dataset}
We identified 8 different entity-entity relation pairs within the n2c2 dataset, and 2 entity-entity relation pairs within the ADE corpus, each with varying instances of the relation pairs (\textbf{Table 1, Figure 3}). Token length distributions of the text, and example of individual entities in n2c2 and ADE dataset are available in  \ref{subsec:tokenlength}, \ref{subsec:examples}.
\subsection{Performance Results}
\noindent \textit{Results on n2c2 and ADE Dataset}\newline 
Our results suggests that integrating prior knowledge from UMLS in the prompts have significant performance improvement as demonstrated by the higher average F-1 scores across both n2c2 and ADE datasets (\textbf{Table 4}).  The reported results are average across all entity-entity relation pairs across models and for 2 datasets. GPT-4-32k model with UMLS show 4\% improvement of F-1 score on the n2c2 dataset, and 12\% improvement on the ADE dataset from the F-1 score of GPT-4-32k model without knowledge integration. For every entity-entity relation pairs, there was a performance improvement by a few percentages for both models and across both datasets. Additional detailed results for each entity-entity relation pairs can be found in Appendices \ref{subsec:n2c2A4} through \ref{subsec:n2c2A9}. \newline 
\indent Upon a closer look at the results, we identified that prompts with UMLS resulted in additional concepts verifying that UMLS is able to identify medications from the text that GPT may not identify independently(Appendix \ref{subsec:results}). \newline 

\noindent \textit{Comparison with Retrieval Augmented Generation}\newline 
\indent RAG model and GPT-3.5-turbo had low F-1 score and it improved with UMLS for both models, but it did not have higher score compared to the GPT-4-32k+UMLS.
\newline
\indent We observed performance variations across entity-entity relation pairs with retrieval augmented generation (\ref{subsec:n2c2A6}, \ref{subsec:n2c2A9}). While some entity pairs showed performance enhancements, others did not show significant improvements. This discrepancy might arise from the limitations of RAG, particularly its inability to utilize entire UMLS thesaurarus in the generation process.  Since UMLS data is partitioned into chunks for indexing and embedding and embedding models can only take 8192 tokens per index, some concepts may not be in the top-k extracted documents used for generation, potentially limiting the scope of augmentation and its impact on final relation pairs. Further experiments are required to confirm this hypothesis.

\section{Discussion and Conclusion}
Our study highlights the significance of merging the strengths of domain-specific knowledge bases, such as UMLS, with the contextual understanding capabilities of LLMs, such as GPT. Our hybrid approach, integrating mapped UMLS concepts with GPT, shows improvement in the model's ability to identify specific entities not inherently within its training data. 

Our results on entity and relation extraction task indicated that leveraging mapped UMLS concepts as additional guidance to the GPT model, helps create focused and unique prompts that significantly enhances GPT's performance. This approach proves more effective than the standard RAG technique. 

In conclusion, the ability to generate tailored prompts based on UMLS concepts offers a promising avenue for improving accuracy and relevance of extracted entities, ultimately enhancing the utility of LLMs in biomedical text analysis tasks.

\section{Limitations and Future Work}
While our framework has shown significant improvements, we acknowledge several limitations in this study. Firstly, our work focused solely on medication concepts, which may restrict the generalizability of our findings to other concepts. However, our approach is adaptable to incorporate additional UMLS entities through prompt adjustments. Future research will explore harnessing UMLS's rich semantic metadata to leverage additional concept relationships, enabling the extraction of a broader spectrum of entity groups beyond medications.

Secondly, our comparison was limited to two generative models, GPT-4-32k and GPT-3.5-turbo. Though they have good performance, we have not included recent models that could have comparable performance. Future work will explore additional models, such as BioGPT, and LAMA for comprehensive comparison and evaluation. This expanded comparison will provide a more nuanced understanding of the performance and capabilities of various generative models in relation to UMLS integration and RAG techniques.\newline 
\indent These future tasks will advance our understanding of the role of domain-specific knowledge in enhancing LLM capabilities and facilitating more effective clinical information extraction. 
\section{Ethics Statement}
IRB approval was not required for this task. To input our text data into the language models, we use Microsoft's Azure OpenAI REST API Service within the Washington University tenant to access OpenAI's language models . We are on a HIPPA-compliant subscription and exempted from content filtering, data review and human review for our use of the Azure OpenAI service. 

\bibliographystyle{acl_natbib}
\bibliography{custom}
 
\appendix
\section{Appendix}
\label{sec:appendix}

\subsection{Token length of the text in (A) n2c2 and (B) ADE dataset}
\label{subsec:tokenlength}
  \begin{figure}[H]
  \centering
  A
  \includegraphics[width=0.5\textwidth]{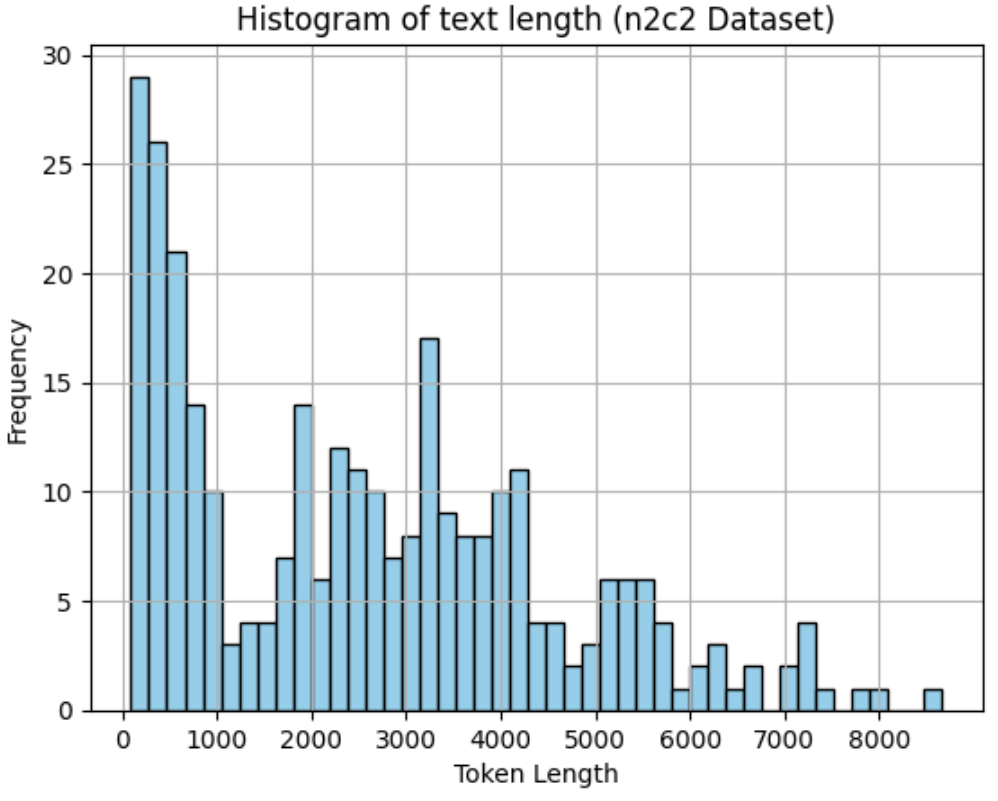}
  B
  \includegraphics[width=0.5\textwidth]{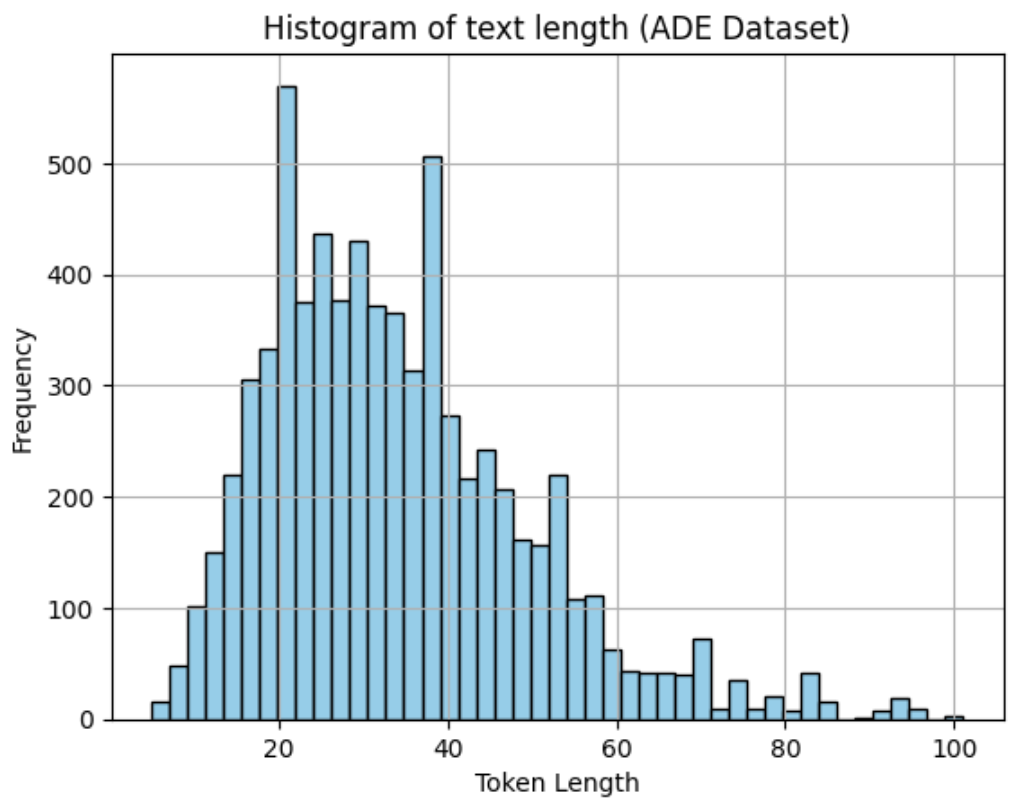}
  \label{fig:example-pdf}
\end{figure}

\subsection{Hyperparameters for the GPT models}
\label{subsec:hyperparameters}
\begin{table} [H]
\centering

\begin{tabular}{p{5cm} p{2cm}} 
\hline
\textbf{Hyperparameters} & \textbf{Value} \\ \hline
Tokenization and Context \newline Window & 200 tokens \\ \hline
Temperature (Randomness of the model output) & 0 \\ \hline
Top p (Top-K Sampling \newline Technique) & 0.95 \\ \hline
Presence Penalty (Penalty to \newline discourage model from \newline generating responses that contain certain specified tokens) & -1.0 \\ \hline
\end{tabular}

\end{table}

\subsection{Example of the individual entities within the n2c2 and ADE dataset}
\label{subsec:examples}
\begin{table}[H]
  \centering
  \caption{Example of the individual entities within the n2c2 and ADE dataset}
  \begin{tabular}{p{2cm} p{5cm}}
    \toprule
    \textbf{Entities} & \textbf{Examples} \\
    \midrule
    Drug & Morphine, ibuprofen, antibiotics (abx), chemotherapy (carboplatin) \\
    ADE/Reason & Nausea, rash, seizures, vitamin K deficiency \\
    Strength & 10 mg, 60 mg \\
    Form & Capsule, syringe, tablet, topical (apply topical) \\
    Dosage & 60 mg/0.6 mL \\
    Frequency & Daily, twice a day, Q4H (every 4 hours) \\
    Route & Transfusion, oral, intravenous (IV) \\
    Duration & For 10 days, 2 cycles, for a week \\
    \bottomrule
  \end{tabular}
  \label{tab:entities}
\end{table}

\newpage

\subsection{Retrieval Augmented Generation}
\label{subsec:RAG}
Method: \newline
1. Split UMLS data (MRCONSO.RRF\footnote{\url{https://www.ncbi.nlm.nih.gov/books/NBK9685/table/ch03.T.concept_names_and_sources_file_mr/}}) into manageable chunks (8192 tokens) to facilitate processing. MRCONSO.RRF file contains the UMLS concepts. \newline
2. Generate embeddings for each chunk, capturing its semantic represetations \newline
2. Store the embeddings in a vector database for efficient retrieval \newline
3. Compare each prompt with the stored data in the vector database. \newline
4. Extract the top 30 results with the highest similarity scores between the query and the UMLS data. \newline
5. Concatenate the retrieved results with the prompt to generate the final extraction output. \newline

\subsection{Qualitative Results}
\label{subsec:results}

\begin{table}[H]
  \caption{Some of the qualitative results for the Strength-Drug Pair. (A) Without UMLS integration. (B) With UMLS integration}
  \begin{tabular}{p{2cm} p{5cm}}
    \toprule
    \textbf{} & \textbf{Examples} \\
    \midrule
    A & [(\textquoteleft aspirin', \textquoteleft 81 mg')
(\textquoteleft atorvastatin', \textquoteleft 20 mg'),
(\textquoteleft amiodarone', \textquoteleft 200 mg'), 
(\textquoteleft metoprolol tartrate', \textquoteleft 50 mg'), 
(\textquoteleft spironolactone', \textquoteleft 25 mg'), 
(\textquoteleft acetaminophen', \textquoteleft 325 mg'), 
(\textquoteleft ranitidine HCl', \textquoteleft 150 mg'), 
(\textquoteleft prednisone', \textquoteleft 60 mg')]
 \\
    \midrule 
    B & [(\textquoteleft aspirin', \textquoteleft 81 mg'), 
(\textquoteleft atorvastatin', \textquoteleft 20 mg'),
 (\textquoteleft amiodarone', \textquoteleft 200 mg'), 
(\textquoteleft metoprolol tartrate', \textquoteleft 50 mg'), 
(\textquoteleft spironolactone', \textquoteleft 25 mg'), 
(\textquoteleft acetaminophen', \textquoteleft 325 mg'), 
(\textquoteleft ranitidine HCl', \textquoteleft 150 mg'), 
(\textquoteleft prednisone', \textquoteleft 60 mg'), 
\textbf{(\textquoteleft Plavix', \textquoteleft 75 mg')},
\textbf{(\textquoteleft ASA', \textquoteleft 325')}, 
\textbf{(\textquoteleft Cipro', \textquoteleft 250 mg')}]
\\
    \bottomrule
  \end{tabular}
  \label{tab:entities}
\end{table}

\clearpage
\subsection{Comparison of GPT-3.5-turbo for all Entity-Entity Relation pairs with and without UMLS Integration for the n2c2 dataset}
\label{subsec:n2c2A4}
\begin{table}[H]
\centering
\label{tab:relation-comparison}
\begin{tabular}{|p{3cm}|ccc|ccc|}
\hline
\multirow{2}{*}{\textbf{Entity-Entity}} & \multicolumn{3}{c|}{\textbf{GPT-3.5-turbo}} & \multicolumn{3}{c|}{\textbf{GPT-3.5-turbo+UMLS}} \\ \cline{2-7} 
& \textbf{P} & \textbf{R} & \textbf{Micro F-1} & \textbf{P} & \textbf{R} & \textbf{F-1} \\ \hline
Dosage-Drug & 0.75 & 0.75 & 0.75 & 0.80 & 0.80 & 0.80 \\
Duration-Drug & 0.76 & 0.76 & 0.76 & 0.81 & 0.81 & 0.81 \\
Route-Drug & 0.74 & 0.73 & 0.73 & 0.76 & 0.75 & 0.75 \\
Form-Drug & 0.72 & 0.73 & 0.72 & 0.75 & 0.76 & 0.75 \\
ADE-Drug & 0.69 & 0.71 & 0.70 & 0.74 & 0.75 & 0.75 \\
Reason-Drug & 0.73 & 0.74 & 0.74 & 0.76 & 0.77 & 0.777 \\
Frequency-Drug & 0.73 & 0.74 & 0.73 & 0.75 & 0.76 & 0.77 \\
\hline
Average & 0.73 & 0.74 & 0.73 & 0.77 & 0.77 & 0.77 \\ \hline
\end{tabular}
\end{table}

\subsection{Comparison of GPT-4-32k for all Entity-Entity Relation pairs without UMLS Integration for the n2c2 dataset}
\label{subsec:n2c2A5}
\begin{table}[H]
\centering
\label{tab:relation-comparison}
\begin{tabular}{|p{3cm}|ccc|ccc|}
\hline
\multirow{2}{*}{\textbf{Entity-Entity}} & \multicolumn{3}{c|}{\textbf{GPT-4-32k}} & \multicolumn{3}{c|}{\textbf{GPT-4-32k+UMLS}} \\ \cline{2-7} 
& \textbf{P} & \textbf{R} & \textbf{Micro F-1} & \textbf{P} & \textbf{R} & \textbf{F-1} \\ \hline
Dosage-Drug & 0.77 & 0.77 & 0.77 & 0.82 & 0.82 & 0.82 \\
Duration-Drug & 0.78 & 0.77 & 0.78 & 0.83 & 0.82 & 0.83 \\
Route-Drug & 0.79 & 0.77 & 0.78 & 0.81 & 0.78 & 0.79 \\
Form-Drug & 0.74 & 0.76 & 0.74 & 0.77 & 0.79 & 0.77 \\
ADE-Drug & 0.69 & 0.73 & 0.71 & 0.75 & 0.78 & 0.77 \\
Reason-Drug & 0.74 & 0.75 & 0.735 & 0.77 & 0.78 & 0.76 \\
Frequency-Drug & 0.78 & 0.77 & 0.78 & 0.80 & 0.79 & 0.79 \\
\hline
Average & 0.75 & 0.76 & 0.76 & 0.79 & 0.79 & 0.79 \\ \hline
\end{tabular}
\end{table}

\subsection{Comparison of Models for all Entity-Entity Relation pairs with UMLS for RAG on the n2c2 dataset}
\label{subsec:n2c2A6}
\begin{minipage}{\linewidth}
\begin{table}[H]

\centering
\label{tab:relation-comparison}
\begin{tabular}{|p{3cm}|ccc|ccc|}
\hline
\multirow{2}{*}{\textbf{Entity-Entity}} & \multicolumn{3}{c|}{\textbf{GPT-4-32k}} & \multicolumn{3}{c|}{\textbf{GPT-3.5-turbo}} \\ \cline{2-7} 
& \textbf{P} & \textbf{R} & \textbf{F-1} & \textbf{P} & \textbf{R} & \textbf{F-1} \\ \hline
Dosage-Drug & 0.77 & 0.77 & 0.77 & 0.75 & 0.75 & 0.75 \\
Duration-Drug & 0.79 & 0.78 & 0.78 & 0.76 & 0.77 & 0.77 \\
Route-Drug & 0.79 & 0.77 & 0.78 & 0.74 & 0.73 & 0.73 \\
Form-Drug & 0.74 & 0.73 & 0.74 & 0.73 & 0.74 & 0.74 \\
ADE-Drug & 0.72 & 0.73 & 0.72 & 0.70 & 0.70 & 0.70 \\
Reason-Drug & 0.76 & 0.76 & 0.76 & 0.75 & 0.78 & 0.76 \\
Frequency-Drug & 0.81 & 0.80 & 0.80 & 0.70 & 0.71 & 0.71 \\
\hline
Average & 0.77 & 0.76 & 0.76 & 0.73 & 0.74 & 0.74 \\ \hline
\end{tabular}
\end{table}
\end{minipage}

\clearpage
\subsection{Comparison of  GPT-4-32k for all Entity-Entity Relation pairs with and without UMLS on the ADE dataset}
\label{subsec:n2c2A7}
\begin{minipage}{\linewidth}
\begin{table}[H]

\centering
\label{tab:relation-comparison}
\begin{tabular}{|p{3cm}|ccc|ccc|}
\hline
\multirow{2}{*}{\textbf{Entity-Entity}} & \multicolumn{3}{c|}{\textbf{GPT-4-32k}} & \multicolumn{3}{c|}{\textbf{GPT-4-32k+UMLS}} \\ \cline{2-7} 
& \textbf{P} & \textbf{R} & \textbf{F-1} & \textbf{P} & \textbf{R} & \textbf{F-1} \\ \hline
Dosage-Drug & 1.0 & 0.66 & 0.795 & 1.00 & 0.85 & 0.91 \\
ADE-Drug & 1.0 & 0.73 & 0.84 & 1.00 & 0.93 & 0.97 \\
\hline
Average & 1.0 & 0.70 & 0.82 & 1.0 & 0.89 & 0.94 \\ \hline
\end{tabular}
\end{table}
\end{minipage}

\subsection{Comparison of  GPT-3.5-turbo for all Entity-Entity Relation pairs with and without UMLS on the ADE dataset}
\label{subsec:n2c2A8}
\begin{minipage}{\linewidth}
\begin{table}[H]

\centering
\label{tab:relation-comparison}
\begin{tabular}{|p{3cm}|ccc|ccc|}
\hline
\multirow{2}{*}{\textbf{Entity-Entity}} & \multicolumn{3}{c|}{\textbf{GPT-3.5-turbo}} & \multicolumn{3}{c|}{\textbf{GPT-3.5-turbo+UMLS}} \\ \cline{2-7} 
& \textbf{P} & \textbf{R} & \textbf{F-1} & \textbf{P} & \textbf{R} & \textbf{F-1} \\ \hline
ADE-Drug & 0.57 & 0.53 & 0.55 & 0.60 & 0.65 & 0.62 \\
Dosage-Drug & 0.68 & 0.61 & 0.64 & 0.70 & 0.75 & 0.72 \\
\hline
Average & 0.625 & 0.57 & 0.596 & 0.83 & 0.70 & 0.75 \\ \hline
\end{tabular}
\end{table}
\end{minipage}

\subsection{Comparison of  the models for all Entity-Entity Relation pairs with UMLS for RAG on the ADE dataset}
\label{subsec:n2c2A9}
\begin{minipage}{\linewidth}
\begin{table}[H]

\centering
\label{tab:relation-comparison}
\begin{tabular}{|p{3cm}|ccc|ccc|}
\hline
\multirow{2}{*}{\textbf{Entity-Entity}} & \multicolumn{3}{c|}{\textbf{GPT-4-32k}} & \multicolumn{3}{c|}{\textbf{GPT-3.5-turbo}} \\ \cline{2-7} 
& \textbf{P} & \textbf{R} & \textbf{F-1} & \textbf{P} & \textbf{R} & \textbf{F-1} \\ \hline
ADE-Drug & 1.0 & 0.73 & 0.84 & 0.62 & 0.61 & 0.60 \\
Dosage-Drug & 1.0 & 0.75 & 0.86 & 0.68 & 0.65 & 0.66 \\
\hline
Average & 1.0 & 0.74 & 0.85 & 0.65 & 0.63 & 0.64 \\ \hline

\end{tabular}
\end{table}
\end{minipage}

\end{document}